\documentclass[a4paper]{article}

\usepackage{INTERSPEECH2022}
\usepackage{tipa}
\usepackage{hyperref}

\title{Large vocabulary speech recognition for languages of Africa: multilingual modeling and self-supervised learning}
\name{Sandy Ritchie, You-Chi Cheng, Mingqing Chen,\\ Rajiv Mathews, Daan van Esch, Bo Li, Khe Chai Sim}
\address{Google Research}
\email{\{sandyritchie,youchicheng,mingqing,mathews,dvanesch,boboli,khechai\}@google.com}

\begin{document}

\maketitle
 
\begin{abstract}
     Almost none of the 2,000+ languages spoken in Africa have widely available automatic speech recognition systems, and the required data is also only available for a few languages. We have experimented with two techniques which may provide pathways to large vocabulary speech recognition for African languages: multilingual modeling and self-supervised learning. We gathered available open source data and collected data for 15 languages, and trained experimental models using these techniques. Our results show that pooling the small amounts of data available in multilingual end-to-end models, and pre-training on unsupervised data can help improve speech recognition quality for many African languages.

\end{abstract}

\section{Introduction}

Africa is home to over 2,000 languages, but almost none of them have widely available automatic speech recognition (ASR) systems. The data required to train such systems is also only available for a few of the major languages, and not in the quantities typically available for high-resource languages. As more people in Africa come online in the 2020s and beyond, the need for high quality ASR systems for African languages becomes more pressing, and along with it the need to develop data sets, model types and training architectures  which are optimized for these languages. 

Two strands in ASR research which may provide solutions are multilingual end-to-end (E2E) modeling \cite{toshniwal2018multilingual,li2021scaling} and self-supervised learning (SSL) \cite{hwang2021large,misra2021comparison}. These techniques have been applied in the low-resource scenario with some encouraging results \cite{cui2015multilingual,dalmia2018sequence,zhou2018multilingual,yi2020applying}, and pre-training  has also been shown to help with accented speech \cite{aksenova2022training}. However, efforts to develop ASR for African languages have so far mostly focused on data collection and model training for single languages or a small number of related or similar languages, including Amharic \cite{abate2005amharic,das2016automatic}; Dinka \cite{das2016automatic}; South African languages \cite{niesler2006language,badenhorst2011collecting,henselmans2013baseline}; Nigerian English \cite{amuda2010limited}; Hausa \cite{schlippe2012hausa,ibrahim2022development}; Yoruba \cite{van2012tone,atanda2013yoruba,adetunmbi2016development,gutkin2020developing}; Swahili \cite{das2016automatic,gelas2012developments,kimutai2013isolated}; Wolof \cite{gauthier2016collecting}; Somali \cite{biswas2019improved}; Igbo \& Fon \cite{dossou2021okwugb}; Bemba \cite{sikasote2021bembaspeech} and Akan \cite{boakye2021research}. 

Alongside the data collections initiated for these projects, there are also a few open source projects which aim to improve African language data coverage. Mozilla Common Voice (MCV)\footnote{\url{https://commonvoice.mozilla.org/}} provides a web-based platform for donating audio data \cite{commonvoice:2020}, and has released data sets for Hausa, Kinyarwanda, Kabyle and Luganda, with Igbo and Swahili coming soon. Notably, and in large part thanks to the efforts of Digital Umuganda, an artificial intelligence company based in Kigali,\footnote{\url{https://digitalumuganda.com/}} MCV provides over 2,000 hours of supervised data for Kinyarwanda, the second largest resource for a language on their platform after English. The South African Centre for Digital Language Resources (SADiLaR)\footnote{\url{https://sadilar.org/}} also provides access to various resources, including the NCHLT speech corpus of the South African languages \cite{barnard2014nchlt}. There are other efforts under the umbrella of the Lacuna Fund\footnote{\url{https://lacunafund.org/}} to create even more data sets, but there is still lots to be done in this space.

\begin{table*}[th]
  \renewcommand\thetable{2}
  \caption{Phonological features and orthographic conventions}
  \label{tab:phon-orth-info}
  \centering
\begin{tabular}{llllll}
\toprule
\textbf{Language} & \textbf{ISO} & \textbf{Tones}     & \textbf{Clicks}    & \textbf{Special characters}         & \textbf{Example word} \\
\midrule
Akan & ak              & high, low          & no                 & open vowels: \textepsilon, \textopeno & akok\textopeno `chicken'       \\
Hausa & ha             & high, low, falling & no                 & implosives: \texthtb, \texthtd, \texthtk, palatalized glottal stop: 'y    & {\texthtb}aataa `spoil'        \\
Igbo & ig              & high, (mid), low   & no                 & \`{V} \'{V} for tone; \d{V} for open vowels & ákw\d{u}kw\d{o} `book'        \\
Ndebele & nr           & high, low          & yes                & none  & umuntu `person'       \\
Northern Sotho & nso              & high, low          & no & š for `sh' sound; \^{V} for open vowels          & mošomô `work'         \\
Kinyarwanda & rw       & high, low          & no                 & none  & urugo `home'          \\
Swati & ss             & high, mid, low     & yes                & none  & sandla `hand'         \\
Southern Sotho & st    & high, low          & yes                & none  & lebese `milk'         \\
Swahili & sw           & none               & no                 & none  & mtoto `child'         \\
Tswana & tn            & high, low          & marginal           & š for `sh' sound; \^{V} for open vowels          & šiti `cloth'          \\
Tsonga & ts            & high, low          & marginal & none    & xitulu `chair'        \\
Venda & ve             & high, low          & no & dentals: 
\textsubcircum{d}, \textsubcircum{l}, \textsubcircum{n}, \textsubcircum{t}; velar \.n        & \textsubcircum{t}un\textsubcircum{d}a `support'     \\
Xhosa & xh             & high, low          & yes                & none  & unxweme `shore'       \\
Yoruba & yo            & high, mid, low     & no & \`{V} \'{V} for tone; \d{s} for `sh' sound; \d{V} for open vowels & k\d{o}já `cross'          \\
Zulu & zu              & high, low, falling & yes                & none  & amanzi `water'       \\
\bottomrule
\end{tabular}
\end{table*}

In general, technology companies have not yet invested significantly in the development of ASR systems for African languages. For a number of years, Google has supported voice search and voice typing for four African languages: Afrikaans, Amharic, Swahili and Zulu,\footnote{\url{https://cloud.google.com/speech-to-text/docs/languages}} but many more languages are spoken in Africa. In this paper, we report on our work to develop or improve ASR systems for 15 African languages: Akan,
Hausa, Igbo, Ndebele, Northern Sotho, Kinyarwanda, Swati, Southern Sotho, Swahili, Tswana, Tsonga, Venda, Xhosa, Yoruba and Zulu. This group of languages was selected for various reasons. Firstly, they are major languages in key regions and countries. Secondly, they share many features in common: they all use the Latin alphabet as their primary writing system, with the exception of Hausa they are all members of the Atlantic-Congo language family, and with the exception of Swahili they are all tonal languages. Finally, some kind of data is available for all of these languages, primarily from open source repositories or through data collections conducted by Google.

Using the available open source and collected data, we trained ASR classic hybrid models, multilingual E2E models, and SSL models. The classic ASR models consist of dedicated connectionist temporal classification (CTC) acoustic models \cite{graves2006connectionist}, pronunciation models and FST-based n-gram language models for each language. We also trained a multilingual Conformer model: a type of LSTM-based recurrent neural network transducer (RNN-T) \cite{he2019streaming} in which the recurrent encoder is replaced with convolution-augmented transformer layers \cite{gulati2020conformer}. For our first experiment with SSL, we pre-trained a model with unsupervised data using wav2vec 2.0 \cite{baevski2020wav2vec} and finetuned it for each of the languages, following \cite{zhang2021bigssl}. We also experimented with pre-training on a larger high-resource unsupervised data set. 

Our results show that pooling the available data in multilingual models, and pre-training on unsupervised data show improvements for African languages compared with classic ASR models. However, while lots of efforts have been spent on creating open source data sets, we have found that the lack of high quality data, which is typically available for high-resource languages, has limited our ability to train models which meet the quality bar for launch in end-user-facing applications.

\section{Languages}

The 15 languages which are the focus of this paper are spoken across West, East and Southern Africa by an estimated 242 million people \cite{vanesch2022writing}. With the exception of Hausa which is an Afroasiatic language, all the languages form part of the large Atlantic-Congo language family. 11 of the 15 are also more closely related in the Bantu group. See Table \ref{tab:lang-info} for the families and population estimates for each language.

\begin{table}[th]
  \renewcommand\thetable{1}
  \caption{Language families and population counts}
  \label{tab:lang-info}
  \centering
\begin{tabular}{llr}
\toprule
\textbf{ISO} & \textbf{Family}     & \multicolumn{1}{l}{\textbf{Population}}  \\
\midrule
ak              & Central Tano          & 13M \\
ha             & Chadic (Afroasiatic)           & 70M \\
ig              & Igboid                   & 18M \\
nr           & Nguni (Bantu)         & 1M  \\
nso              & Sotho-Tswana (Bantu)  & 14M \\
rw       & Rwanda-Rundi (Bantu)      & 11M \\
ss             & Nguni (Bantu)          & 3M  \\
st    & Sotho-Tswana (Bantu)         & 14M \\
sw           & Sabaki (Bantu)                 & 24M \\
tn            & Sotho-Tswana (Bantu)   & 8M  \\
ts            & Tswa-Ronga (Bantu)         & 13M \\
ve             & Southern Bantu        & 2M  \\
xh             & Nguni (Bantu)         & 19M \\
yo            & Yoruboid             & 21M \\
zu              & Nguni (Bantu)         & 11M \\
\bottomrule
\end{tabular}
\end{table}

We include Hausa in our group despite its different lineage because it shares some common features with its Atlantic-Congo neighbours, most notably lexical and grammatical tone. In grammatical tone systems, each word has an inherent tonal pattern, and these patterns can also change to indicate grammatical features, for example tense and aspect on verbs \cite{odden1995tone}. All languages in the group except for Swahili exhibit lexical and/or grammatical tone. Another distinctive phonological feature common among several of the languages is the use of click consonants. These are especially prevalent in the Southern Bantu languages, and likely came into these languages through contact with neighbouring Khoisan languages \cite{herbert1990sociohistory}. Xhosa has a particularly large set of 18 click consonants.

All the languages use a form of the Latin script as their primary writing system. In some languages like Xhosa and Zulu, no special characters are used, and phonological features like clicks are represented using Latin letters like $<$c$>$ and $<$x$>$. In other languages, the basic Latin set is supplemented with other characters and diacritics to represent tone and other phonological features. Commonly, high tone is marked with an acute accent, and low tone is marked with a grave accent on tone-bearing vowel and nasal graphemes. Less commonly, other characters like macron, circumflex or caron are used to mark mid tone \cite{bird1999strategies}. Some languages like Hausa do not mark tone in the orthography. A summary of some phonological features and orthographic conventions used for the 15 languages is given in Table \ref{tab:phon-orth-info}.\footnote{ISO 639 codes from \url{https://iso639-3.sil.org/}.}

\section{Data}
\label{sec:data}

In Tables \ref{tab:supervised-data} and \ref{tab:unsup-data}, the volume of supervised and unsupervised data we used for each language is given. We have used available open source data where possible, for example we have a million utterances and 1,406 hours of training data for Kinyarwanda thanks mostly to the open source repository provided by MCV,\footnote{The Kinyarwanda MCV corpus has around 2,000 hours of data, of which we use around 1,000 hours for training and 300 hours for testing. The rest of the data was not used due to issues with alignment.} and for the South African languages, we rely almost entirely on the NCHLT speech corpus provided through SADiLaR; we only have supplementary data for Southern Sotho and Zulu. We split off approximately 20\% of the original data set in each case and used that for evaluation of the various models.

\begin{table}[th]
  \caption{Supervised data - utterances and audio hours \\(Only NCHLT data for most South African languages)}
  \label{tab:supervised-data}
  \centering
\begin{tabular}{lrrrr}
\toprule
\textbf{ISO} & \multicolumn{2}{c}{\textbf{Short form}}  & \multicolumn{2}{c}{\textbf{Long form}}  \\
\textbf{}   & \multicolumn{1}{c}{utts} & \multicolumn{1}{c}{hrs} & \multicolumn{1}{c}{utts} & \multicolumn{1}{c}{hrs}     \\
\midrule
ak & 247K & 253.07 & \multicolumn{1}{r}{1514} & \multicolumn{1}{r}{193.60}  \\
ha  & 281K & 169.38 & \multicolumn{1}{r}{1757} & \multicolumn{1}{r}{193.20}  \\
ig & 262K & 144.73 & \multicolumn{1}{r}{2417} & \multicolumn{1}{r}{347.80}  \\
nr & 39K  & 51.58 & -- &  -- \\
nso & 56K  & 53.41 & -- &  -- \\
rw & 1M & 1405.75  & \multicolumn{1}{r}{952}  & \multicolumn{1}{r}{156.90}  \\
ss & 41K  & 53.00 & -- &  -- \\
st  & 191K & 154.23 & -- &  -- \\
sw & 515K & 572.65 & \multicolumn{1}{r}{4137} & \multicolumn{1}{r}{1519.56} \\
tn & 56K  & 53.57 & -- &  -- \\
ts & 45K  & 52.66 & -- &  -- \\
ve  & 47K  & 53.15 & -- &  -- \\
xh  & 44K  & 53.15 & -- &  -- \\
yo & 259K & 139.61 & \multicolumn{1}{r}{1426} & \multicolumn{1}{r}{271.10}  \\
zu & 673K & 1011.04  & \multicolumn{1}{r}{1659} & \multicolumn{1}{r}{315.50}  \\
\midrule
TOTAL & 3.8M & 4220.98 & 13862 & 2997.66 \\
    \bottomrule
  \end{tabular}
  
\end{table}

\begin{table}[th]
  \caption{Unsupervised data - utterances and audio hours \\(Only NCHLT data for most South African languages, no short form data for currently unsupported languages)}
  \label{tab:unsup-data}
  \centering
\begin{tabular}{lrrrr}
\toprule
\textbf{ISO} & \multicolumn{1}{c}{\textbf{Short form}}  & \multicolumn{2}{c}{\textbf{Long form}}  \\
\textbf{}   & \multicolumn{1}{c}{utts} & \multicolumn{1}{c}{utts} & \multicolumn{1}{c}{hrs}     \\
\midrule
ak & \multicolumn{1}{c}{--} & 19K & 2794.41\\
ha  & \multicolumn{1}{c}{--} & 50K & 8230.37 \\
ig & \multicolumn{1}{c}{--} & 2949 & \multicolumn{1}{r}{2366.77} \\
nr  & \multicolumn{1}{c}{--} & --  & --\\
nso & \multicolumn{1}{c}{--} & --  & --\\
rw & \multicolumn{1}{c}{--} & 654  & \multicolumn{1}{r}{239.21}  \\
ss  & \multicolumn{1}{c}{--} & --  & --\\
st  & \multicolumn{1}{c}{--} & 281  & \multicolumn{1}{r}{137.60}  \\
sw  & \multicolumn{1}{c}{96K} & --  & --\\
tn & \multicolumn{1}{c}{--} & -- & --\\
ts & \multicolumn{1}{c}{--} & --  & --\\
ve  & \multicolumn{1}{c}{--} & --  & --\\
xh  & \multicolumn{1}{c}{--} & 128  & \multicolumn{1}{r}{36.35}   \\
yo & \multicolumn{1}{c}{--} & 6015 & \multicolumn{1}{r}{3578.37} \\
zu & \multicolumn{1}{c}{39K} & 107  & \multicolumn{1}{r}{34.44} \\
\midrule
TOTAL & \multicolumn{1}{c}{135K} & 79375 & 17417.53 \\
    \bottomrule
  \end{tabular}
  
\end{table}

The supervised short form data was collected in two ways. First is prompt-based collection, where contributors are shown written sentences (i.e. prompts) and record themselves reading them out loud. In the second type of collection, contributors are shown images of common objects and scenes and record themselves describing the image in a few words. These descriptions are then transcribed by other contributors. Supervised long form data was collected by identifying public videos on YouTube in the target language and transcribing them. The videos were manually identified by linguists at Google and verified as containing content in the target language by the transcribers. We applied the voice activity detection model used in \cite{zhang2021bigssl} to segment the video.

The unsupervised short form data was collected from voice search queries where users had opted in to help Google develop and improve its audio recognition technologies and the Google services that use them.

The unsupervised long form data was collected by identifying public YouTube videos and extracting the audio from the video.

\section{Experiments}

We conducted our experiments with four settings. The details are as follows:

(1) \textbf{Classic ASR models}. We developed grapheme-to-phoneme (G2P) rules to generate pronunciation lexicons, gathered text data and developed text normalization grammars to train n-gram language models, and trained custom CTC acoustic models for seven of the languages. We did not train classic models for Ndebele, Northern Sotho, Swati, Southern Sotho, Tswana, Tsonga, Venda, Xhosa or Yoruba, because at the time the NCHLT corpus was not available to us, and in the case of Yoruba because we had issues with data quality.

(2) \textbf{Multilingual Conformer}. We trained a multilingual Conformer model -- this is a modified LSTM-based RNN-T architecture in which the recurrent encoder is replaced with convolution-augmented transformer layers. The encoder is composed of 12 layers, with left 3 frames stacked 128-channel log-mel features for the 512 conformer model dimension. 
A streaming encoder is used, where each local self-attention layer looks at 23 left context and 0 right context. The model is trained with FastEmit \cite{yu2021fastemit} and Hybrid Autoregressive Transducer (HAT) factorization \cite{variani2020hybrid}, where the latter technique enables better integration with an external language model.\footnote{In this work, no external LM is used for all E2E settings.} The prediction network uses a small embedding lookup architecture introduced in \cite{sainath2021efficient}.

(3) \textbf{Multilingual Pre-Training (MPT)}: Following Zhang et al.'s work on BigSSL \cite{zhang2021bigssl}, the unsupervised data listed in Table \ref{tab:unsup-data} are segmented for pre-training with the wav2vec 2.0 objective \cite{baevski2020wav2vec} and are finetuned on each of the 15  languages using the standard RNN-T loss as the downstream task. We used the 600M-parameter Conformer model, the same architecture reported in \cite{zhang2021bigssl} for both pre-training and finetuning.

(4) \textbf{Larger Dataset Multilingual Pre-Training (LDMPT)}: This setting is similar to (3). The difference is that before we pre-trained using unsupervised data from the 15  languages, we first applied the same pre-training objective on a much larger high-resource language unsupervised data set. More specifically, we used 900,000 hours of segmented, unlabeled YT-U audio data from the set reported in \cite{zhang2021bigssl}. The underlying assumption is that this would make the model generalize better, as the model sees more data with different phonetic distribution, even though there is a language mismatch between the high-resource languages and the African languages.


\section{Evaluation}

We evaluated the different model types on test sets composed of supervised short form utterances, which were split off from the acquired or collected data sets (see Section \ref{sec:data}). Table \ref{tab:wers} shows word error rates (WERs) for each type of model and setting.

\begin{table}[th]
  \caption{Word error rates on different training settings}
  \label{tab:wers}
  \centering
  \begin{tabular}{lrrrr}
    \toprule
    {\textbf{ISO}} & \multicolumn{1}{l}{\textbf{Classic}} &  \multicolumn{1}{l}{\textbf{Conformer}} & \multicolumn{1}{l}{\textbf{MPT}} & \multicolumn{1}{l}{\textbf{LDMPT}}\\
    \midrule
    ak & 39.0  & \textbf{26.5} & 44.9 & 44.9 \\
    ha & 37.9  & 32.6 & \textbf{29.7} & 29.9  \\
    ig & 48.4  & \textbf{35.0} & 63.4 & 62.3 \\
    nr &--& 13.3 & 7.4 & \textbf{6.7} \\
    nso &--& \textbf{8.5} & 10.3 & 9.8 \\
    rw & 37.2  & 13.8 & 10.4 & \textbf{9.8} \\
    ss &--& 7.0 & 4.4 & \textbf{3.8} \\
    st &--& 9.1 & \textbf{4.7}  & 5.4 \\
    sw & 23.7 & 23.7 & 6.9 & \textbf{6.4} \\
    tn &--& 4.8 & 1.9 & \textbf{1.4} \\
    ts &--& 7.4 & \textbf{2.3} & \textbf{2.3} \\
    ve &--& 13.8 & 11.0 & \textbf{9.9} \\
    xh &--& 14.2 & 11.1 & \textbf{7.1} \\
    yo &--&  \textbf{45.6} & 62.4 & 60.1 \\
    zu & 23.8 &  23.8 & 15.0 & \textbf{13.1} \\
    \bottomrule
  \end{tabular}
  
\end{table}

\section{Discussion}

In general, the single-language classic ASR models have higher WERs compared with other techniques. This is likely due to the comparatively small amounts of training data available for each individual language. Issues with transcription quality, auto-generated G2P accuracy and scarce text data may also contribute to the higher WERs for classic models. Swahili and Zulu do not exhibit the same disparities as the other languages, likely because these languages have more training data and their lexicons and language models have been subject to greater attention, as Google has supported these two languages for a number of years.

While we cannot draw direct comparisons between the Conformer and pre-training results, as the model types and sizes are quite different, there are some noteworthy patterns in the results. In general, the results for the pre-trained models are better than the Conformer model for the South African languages, where our only source of training data is the NCHLT corpus.\footnote{One counter-example in this trend is Northern Sotho (nso), which shows better results in the Conformer model.} Conversely, where our only source of short-form training data comes from image description tasks, as is the case for Akan, Hausa, Igbo and Yoruba, the Conformer model outperforms the pre-trained models, except in the case of Hausa. In cases where we have both an image description set and another train set, as is the case for Kinyarwanda and Southern Sotho, the pre-trained models show improvements over the Conformer model. 

Comparing the two pre-training settings, we note that 7 languages show improvements if we first pre-train the model with high resource-language data rather than training solely on unsupervised data. This might be due to the fact that the unsupervised data is of inferior quality: many of the videos have background music and might also suffer from audio clipping issues. Potential solutions might be to filter out the speech-music mixed segments and apply some declipping preprocessing algorithms \cite{harvilla2014least}.

In all settings, WERs are higher for languages with only image description data. This suggests two things: (1) relying on this type of data alone is more likely to result in poorer quality models, and (2) the multilingual Conformer model seems to be more resilient to less reliable data.

Looking into the issues with the image description data sets further, we see two types of problems. The first is that some of the utterances are one word in length, and this may present an issue when these utterances are seen in training. The second issue is that there are more discrepancies between the audio and transcriptions than we see with prompt-based corpora, which can be attributed to several factors: spontaneous speech contains more hesitations and other features which are hard to represent accurately in writing, and there is greater potential for transcribers to mishear or misunderstand the audio content.

Issues with transcription quality are not necessarily limited to the image-based data sets. In all the data sets, there are issues with spelling variation and variation in the use of diacritics and special characters outlined in Table \ref{tab:phon-orth-info}. We see the following kinds of variation: (1) alternative spellings for words; (2) alternative symbols instead of the standard ones, typically because the standard characters are not easily accessible on many keyboards or other input methods; (3) use of alternative diacritics, for example carons instead of circumflexes or vice versa; (4) variation in use of tone-marking diacritics: in some languages, while the standard orthography requires marking tone on every vowel and nasal, in practice tone marking is only used in cases where the meaning would otherwise be ambiguous, or tone is not marked at all. While the first three types of variation can be comparatively easily standardized with text normalization \cite{zupon2021text}, deficient or non-existent tone marking is much harder to restore, as the tones associated with words can vary based on grammatical features, meaning that a contextual model is required for tone restoration, e.g. \cite{asahiah2017restoring}. For languages where tone is marked inconsistently, this may play a significant role in the higher WERs we see for some of the 15 languages which are the subject of this paper.



\section{Conclusions}

We have shown that multilingual RNN-T models and self-supervised pre-training techniques can improve ASR quality for African languages. These are just two techniques among many that have been shown to be useful in the low-resource scenario. Other novel ASR modeling techniques which could help include federated learning and personalization, zero-shot learning, data augmentation using synthesized speech (though text-to-speech is generally not available for these languages either), and adding external LMs which would also enable techniques like second pass rescoring. We hope that this work stimulates further research on ASR for these and other languages of Africa.

\section{Acknowledgements}

We would like to thank Parisa Hagani, Manasa Prasad, Isabel Leal, Neeraj Gaur, Brian Farris, Yun Zhu, Al\"{e}na Aks\"{e}nova, Pierric Sans, Landis Baker, Mandy Jordan, Eoin Mahon, Clara Rivera and Wei Han for support and feedback on the work presented in this paper, and Françoise Beaufays and Pedro Moreno for executive support.

\bibliographystyle{IEEEbib}
\bibliography{mybib}

\begin{thebibliography}{10}
\providecommand{\url}[1]{#1}
\csname url@samestyle\endcsname
\providecommand{\newblock}{\relax}
\providecommand{\bibinfo}[2]{#2}
\providecommand{\BIBentrySTDinterwordspacing}{\spaceskip=0pt\relax}
\providecommand{\BIBentryALTinterwordstretchfactor}{4}
\providecommand{\BIBentryALTinterwordspacing}{\spaceskip=\fontdimen2\font plus
\BIBentryALTinterwordstretchfactor\fontdimen3\font minus
  \fontdimen4\font\relax}
\providecommand{\BIBforeignlanguage}[2]{{%
\expandafter\ifx\csname l@#1\endcsname\relax
\typeout{** WARNING: IEEEtran.bst: No hyphenation pattern has been}%
\typeout{** loaded for the language `#1'. Using the pattern for}%
\typeout{** the default language instead.}%
\else
\language=\csname l@#1\endcsname
\fi
#2}}
\providecommand{\BIBdecl}{\relax}
\BIBdecl

\bibitem{toshniwal2018multilingual}
S.~Toshniwal, T.~N. Sainath, R.~J. Weiss, B.~Li, P.~Moreno, E.~Weinstein, and
  K.~Rao, ``Multilingual speech recognition with a single end-to-end model,''
  in \emph{ICASSP 2018}.\hskip 1em plus 0.5em minus 0.4em\relax IEEE, 2018, pp.
  4904--4908.

\bibitem{li2021scaling}
B.~Li, R.~Pang, T.~N. Sainath, A.~Gulati, Y.~Zhang, J.~Qin, P.~Haghani, W.~R.
  Huang, M.~Ma, and J.~Bai, ``{Scaling end-to-end models for large-scale
  multilingual ASR},'' \emph{arXiv preprint arXiv:2104.14830}, 2021.

\bibitem{hwang2021large}
D.~Hwang, A.~Misra, Z.~Huo, N.~Siddhartha, S.~Garg, D.~Qiu, K.~C. Sim,
  T.~Strohman, F.~Beaufays, and Y.~He, ``{Large-scale ASR Domain Adaptation
  using Self-and Semi-supervised Learning},'' \emph{arXiv preprint
  arXiv:2110.00165}, 2021.

\bibitem{misra2021comparison}
A.~Misra, D.~Hwang, Z.~Huo, S.~Garg, N.~Siddhartha, A.~Narayanan, and K.~C.
  Sim, ``{A comparison of supervised and unsupervised pre-training of
  end-to-end models},'' in \emph{INTERSPEECH}, vol. 2021, 2021, pp. 731--735.

\bibitem{cui2015multilingual}
J.~Cui, B.~Kingsbury, B.~Ramabhadran, A.~Sethy, K.~Audhkhasi, X.~Cui,
  E.~Kislal, L.~Mangu, M.~Nussbaum-Thom, M.~Picheny \emph{et~al.},
  ``Multilingual representations for low resource speech recognition and
  keyword search,'' in \emph{2015 IEEE workshop on automatic speech recognition
  and understanding (ASRU)}.\hskip 1em plus 0.5em minus 0.4em\relax IEEE, 2015,
  pp. 259--266.

\bibitem{dalmia2018sequence}
S.~Dalmia, R.~Sanabria, F.~Metze, and A.~W. Black, ``Sequence-based
  multi-lingual low resource speech recognition,'' in \emph{ICASSP 2018}.\hskip
  1em plus 0.5em minus 0.4em\relax IEEE, 2018, pp. 4909--4913.

\bibitem{zhou2018multilingual}
S.~Zhou, S.~Xu, and B.~Xu, ``Multilingual end-to-end speech recognition with a
  single transformer on low-resource languages,'' \emph{arXiv preprint
  arXiv:1806.05059}, 2018.

\bibitem{yi2020applying}
C.~Yi, J.~Wang, N.~Cheng, S.~Zhou, and B.~Xu, ``Applying wav2vec 2.0 to speech
  recognition in various low-resource languages,'' \emph{arXiv preprint
  arXiv:2012.12121}, 2020.

\bibitem{aksenova2022training}
A.~Aks\"{e}nova, Z.~Chen, C.-C. Chiu, D.~van Esch, P.~Golik, W.~Han, L.~King,
  B.~Ramabhadran, A.~Rosenberg, S.~Schwartz, and G.~Wang, ``{Accented speech
  recognition: benchmarking, pre-training, and diverse data},'' 2022.

\bibitem{abate2005amharic}
S.~T. Abate, W.~Menzel, and B.~Tafila, ``{An Amharic speech corpus for large
  vocabulary continuous speech recognition},'' in \emph{Ninth European
  Conference on Speech Communication and Technology}, 2005.

\bibitem{das2016automatic}
A.~Das, P.~Jyothi, and M.~Hasegawa-Johnson, ``{Automatic Speech Recognition
  Using Probabilistic Transcriptions in Swahili, Amharic, and Dinka},'' in
  \emph{INTERSPEECH}, 2016, pp. 3524--3528.

\bibitem{niesler2006language}
T.~Niesler, ``{Language-dependent state clustering for multilingual speech
  recognition in Afrikaans, South African English, Xhosa and Zulu},'' in
  \emph{Multilingual Speech and Language Processing}, 2006.

\bibitem{badenhorst2011collecting}
J.~Badenhorst, C.~Van~Heerden, M.~Davel, and E.~Barnard, ``{Collecting and
  evaluating speech recognition corpora for 11 South African languages},''
  \emph{Language resources and evaluation}, vol.~45, no.~3, pp. 289--309, 2011.

\bibitem{henselmans2013baseline}
D.~Henselmans, T.~Niesler, and D.~Van~Leeuwen, ``{Baseline speech recognition
  of South African languages using Lwazi and AST},'' \emph{Proc. PRASA,
  Johannesburg, South Africa}, pp. 30--35, 2013.

\bibitem{amuda2010limited}
S.~Amuda, H.~Bo{\v{r}}il, A.~Sangwan, and J.~H. Hansen, ``{Limited resource
  speech recognition for Nigerian English},'' in \emph{ICASSP 2010}.\hskip 1em
  plus 0.5em minus 0.4em\relax IEEE, 2010, pp. 5090--5093.

\bibitem{schlippe2012hausa}
T.~Schlippe, E.~G.~K. Djomgang, N.~T. Vu, S.~Ochs, and T.~Schultz, ``{Hausa
  large vocabulary continuous speech recognition},'' in \emph{Spoken Language
  Technologies for Under-Resourced Languages}, 2012.

\bibitem{ibrahim2022development}
U.~A. Ibrahim, M.~M. Boukar, and M.~A. Suleiman, ``{Development of Hausa
  dataset: a baseline for speech recognition},'' \emph{Data in Brief}, vol.~40,
  2022.

\bibitem{van2012tone}
D.~Van~Niekerk and E.~Barnard, ``{Tone realisation in a Yor{\`u}b{\'a} speech
  recognition corpus},'' 2012.

\bibitem{atanda2013yoruba}
A.~Atanda, S.~Yusof, and M.~Hariharan, ``{Yor{\`u}b{\'a} automatic speech
  recognition: A review},'' in \emph{Rural ICT Development (RICTD)
  International Conference}, vol.~1, no.~1, 2013, pp. 116--121.

\bibitem{adetunmbi2016development}
O.~Adetunmbi, O.~Obe, and J.~Iyanda, ``{Development of Standard Yor{\`u}b{\'a}
  speech-to-text system using HTK},'' \emph{International Journal of Speech
  Technology}, vol.~19, no.~4, pp. 929--944, 2016.

\bibitem{gutkin2020developing}
A.~Gutkin, I.~Demirsahin, O.~Kjartansson, C.~E. Rivera, and
  K.~T{\'u}b{\`o}s{\'u}n, ``{Developing an open-source corpus of Yoruba
  speech},'' 2020.

\bibitem{gelas2012developments}
H.~Gelas, L.~Besacier, and F.~Pellegrino, ``{Developments of Swahili resources
  for an automatic speech recognition system},'' in \emph{Spoken Language
  Technologies for Under-Resourced Languages}, 2012.

\bibitem{kimutai2013isolated}
S.~K. Kimutai, E.~Milgo, and D.~Gichoya, ``{Isolated Swahili words recognition
  using Sphinx4},'' \emph{International Journal of Emerging Science and
  Engineering (IJESE)}, vol.~2, no.~2, pp. 2319--6378, 2013.

\bibitem{gauthier2016collecting}
E.~Gauthier, L.~Besacier, S.~Voisin, M.~Melese, and U.~P. Elingui,
  ``{Collecting resources in Sub-Saharan African languages for automatic speech
  recognition: a case study of Wolof},'' in \emph{10th Language Resources and
  Evaluation Conference (LREC 2016)}, 2016.

\bibitem{biswas2019improved}
A.~Biswas, R.~Menon, E.~van~der Westhuizen, and T.~Niesler, ``{Improved
  low-resource Somali speech recognition by semi-supervised acoustic and
  language model training},'' \emph{arXiv preprint arXiv:1907.03064}, 2019.

\bibitem{dossou2021okwugb}
B.~F. Dossou and C.~C. Emezue, ``{OkwuGb$\backslash$'e: End-to-End Speech
  Recognition for Fon and Igbo},'' \emph{arXiv preprint arXiv:2103.07762},
  2021.

\bibitem{sikasote2021bembaspeech}
C.~Sikasote and A.~Anastasopoulos, ``{BembaSpeech: A Speech Recognition Corpus
  for the Bemba Language},'' \emph{arXiv preprint arXiv:2102.04889}, 2021.

\bibitem{boakye2021research}
A.~A. Boakye-Yiadom, M.~Qin, and R.~Jing, ``{Research of Automatic Speech
  Recognition of Asante-Twi Dialect For Translation},'' in \emph{Proceedings of
  the 2021 5th International Conference on Electronic Information Technology
  and Computer Engineering}, 2021, pp. 1086--1094.

\bibitem{commonvoice:2020}
R.~Ardila, M.~Branson, K.~Davis, M.~Henretty, M.~Kohler, J.~Meyer, R.~Morais,
  L.~Saunders, F.~M. Tyers, and G.~Weber, ``Common voice: A
  massively-multilingual speech corpus,'' in \emph{Proceedings of the 12th
  Conference on Language Resources and Evaluation (LREC 2020)}, 2020, pp.
  4211--4215.

\bibitem{barnard2014nchlt}
E.~Barnard, M.~H. Davel, C.~van Heerden, F.~De~Wet, and J.~Badenhorst, ``{The
  NCHLT speech corpus of the South African languages},'' in \emph{Workshop
  Spoken Language Technologies for Under-resourced Languages (SLTU)}, 2014.

\bibitem{graves2006connectionist}
A.~Graves, S.~Fern{\'a}ndez, F.~Gomez, and J.~Schmidhuber, ``Connectionist
  temporal classification: labelling unsegmented sequence data with recurrent
  neural networks,'' in \emph{Proceedings of the 23rd International Conference
  on Machine Learning}, 2006, pp. 369--376.

\bibitem{he2019streaming}
Y.~He, T.~N. Sainath, R.~Prabhavalkar, I.~McGraw, R.~Alvarez, D.~Zhao,
  D.~Rybach, A.~Kannan, Y.~Wu, R.~Pang \emph{et~al.}, ``Streaming end-to-end
  speech recognition for mobile devices,'' in \emph{ICASSP 2019}.\hskip 1em
  plus 0.5em minus 0.4em\relax IEEE, 2019, pp. 6381--6385.

\bibitem{gulati2020conformer}
A.~Gulati, J.~Qin, C.-C. Chiu, N.~Parmar, Y.~Zhang, J.~Yu, W.~Han, S.~Wang,
  Z.~Zhang, Y.~Wu \emph{et~al.}, ``{Conformer: Convolution-augmented
  transformer for speech recognition},'' \emph{arXiv preprint
  arXiv:2005.08100}, 2020.

\bibitem{baevski2020wav2vec}
A.~Baevski, Y.~Zhou, A.~Mohamed, and M.~Auli, ``wav2vec 2.0: A framework for
  self-supervised learning of speech representations,'' \emph{Advances in
  Neural Information Processing Systems}, vol.~33, pp. 12\,449--12\,460, 2020.

\bibitem{zhang2021bigssl}
Y.~Zhang, D.~S. Park, W.~Han, J.~Qin, A.~Gulati, J.~Shor, A.~Jansen, Y.~Xu,
  Y.~Huang, S.~Wang \emph{et~al.}, ``{BigSSL: Exploring the frontier of
  large-scale semi-supervised learning for automatic speech recognition},''
  \emph{arXiv preprint arXiv:2109.13226}, 2021.

\bibitem{vanesch2022writing}
D.~van Esch, T.~Lucassen, S.~Ruder, I.~Caswell, and C.~Rivera, ``{Writing
  System and Speaker Metadata for 2,800+ Language Varieties},'' in
  \emph{Proceedings of LREC}, 2022.

\bibitem{odden1995tone}
D.~Odden, ``{Tone: African languages},'' \emph{The Handbook of Phonological
  Theory}, vol.~1, pp. 444--75, 1995.

\bibitem{herbert1990sociohistory}
R.~K. Herbert, ``{The sociohistory of clicks in Southern Bantu},''
  \emph{Anthropological linguistics}, pp. 295--315, 1990.

\bibitem{bird1999strategies}
S.~Bird, ``{Strategies for representing tone in African writing systems},''
  \emph{Written Language \& Literacy}, vol.~2, no.~1, pp. 1--44, 1999.

\bibitem{yu2021fastemit}
J.~Yu, C.-C. Chiu, B.~Li, S.-y. Chang, T.~N. Sainath, Y.~He, A.~Narayanan,
  W.~Han, A.~Gulati, Y.~Wu \emph{et~al.}, ``{Fastemit: Low-latency streaming
  ASR with sequence-level emission regularization},'' in \emph{ICASSP
  2021}.\hskip 1em plus 0.5em minus 0.4em\relax IEEE, 2021, pp. 6004--6008.

\bibitem{variani2020hybrid}
E.~Variani, D.~Rybach, C.~Allauzen, and M.~Riley, ``{Hybrid autoregressive
  transducer (HAT)},'' in \emph{ICASSP 2020}.\hskip 1em plus 0.5em minus
  0.4em\relax IEEE, 2020, pp. 6139--6143.

\bibitem{sainath2021efficient}
T.~N. Sainath, Y.~R. He, A.~Narayanan, R.~Botros, R.~Pang, D.~J. Rybach,
  C.~Allauzen, E.~Variani, J.~Qin, A.~Gruenstein \emph{et~al.}, ``An efficient
  streaming non-recurrent on-device end-to-end model with improvements to
  rare-word modeling,'' in \emph{INTERSPEECH}, 2021, pp. 1777--1781.

\bibitem{harvilla2014least}
M.~J. Harvilla and R.~M. Stern, ``{Least squares signal declipping for robust
  speech recognition},'' in \emph{INTERSPEECH}, 2014.

\bibitem{zupon2021text}
A.~Zupon, E.~Crew, and S.~Ritchie, ``{Text normalization for low-resource
  languages of Africa},'' \emph{arXiv preprint arXiv:2103.15845}, 2021.

\bibitem{asahiah2017restoring}
F.~O. Asahiah, O.~A. Odejobi, and E.~R. Adagunodo, ``{Restoring tone-marks in
  standard Yor{\`u}b{\'a} electronic text: improved model},'' \emph{Computer
  Science}, vol.~18, no.~3, 2017.

\end{thebibliography}

\end{document}